\documentclass[runningheads]{llncs}
\usepackage{graphicx}

\usepackage{tikz}
\usepackage{comment}
\usepackage{amsmath,amssymb} 
\usepackage{color}

\usepackage[accsupp]{axessibility}  

\usepackage{multirow}
\usepackage{subfigure}

\begin{document}
\pagestyle{headings}
\mainmatter
\def\ECCVSubNumber{2423}  

\title{Explaining Deepfake Detection by Analysing Image Matching} 


\titlerunning{Explaining Deepfake Detection by Analysing Image Matching}
%

{
	\renewcommand{\thefootnote}{\fnsymbol{footnote}}
	
	\author{Shichao Dong$^{1,*}$ \and
		Jin Wang$^{1,*}$ \and
		Jiajun Liang$^{1}$ \and
		Haoqiang Fan$^{1}$ \and
		Renhe Ji$^{1,\dag}$}
	\institute{$^1$MEGVII Technology
		\email{\{dongshichao,wangjin,liangjiajun,fanhaoqiang,jirenhe\}@megvii.com }\\
	}
	\footnotetext[1]{Equal contribution}
	\footnotetext[4]{Corresponding author}
}
\authorrunning{S. Dong et al.}
\maketitle

\begin{abstract}
This paper aims to interpret how deepfake detection models learn artifact features of images when just supervised by binary labels. 
To this end, three hypotheses from the perspective of image matching are proposed as follows. 
1. Deepfake detection models indicate real/fake images based on visual concepts that are neither source-relevant nor target-relevant, that is, considering such visual concepts as artifact-relevant.\
2. Besides the supervision of binary labels, deepfake detection models implicitly learn artifact-relevant visual concepts through the FST-Matching (\emph{i.e.} the matching \textbf{f}ake, \textbf{s}ource, \textbf{t}arget images) in the training set.\
3. Implicitly learned artifact visual concepts through the FST-Matching in the raw training set are vulnerable to video compression. \
In experiments, the above hypotheses are verified among various DNNs.
Furthermore, based on this understanding, we propose the FST-Matching Deepfake Detection Model to boost the performance of forgery detection on compressed videos.
Experiment results show that our method achieves great performance, especially on highly-compressed (\emph{e.g.} c40) videos.
\end{abstract}

\section{Introduction}
Recently, deepfake methods \cite{deepfakes,faceshifter,faceswap,face2face,neural-textural} have exhibited superior performance in synthesizing realistic faces. 
Such face forgeries may easily be used by attackers for malicious purposes, causing severe social problems and political threats. 
To this end, plenty of studies \cite{ff++,mesonet} have achieved great success in detecting various manipulated media by simply considering it as a binary classification task.
However, understanding how these models learn artifact features of images when just supervised by binary labels (real/fake) is still a challenge to state-of-the-art algorithms. 

\begin{figure}[t]
\centering
\includegraphics[height=4cm]{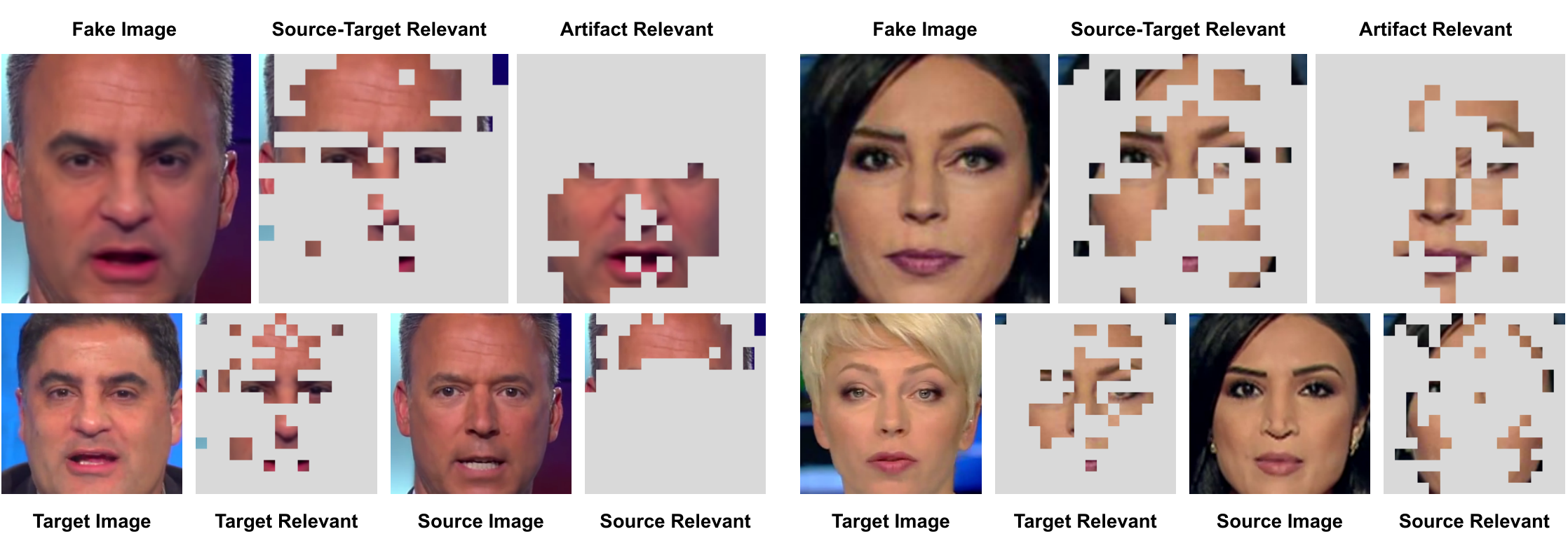}

\caption{\textbf{The relationship between source/target-relevant visual concepts and artifact-relevant visual concepts.} Here, visual concepts represent image regions such as eyes, mouths and foreheads of human faces. In this paper, we find that well-trained deepfake detection models mainly consider artifact-relevant visual concepts as neither source-relevant nor target-relevant from the perspective of image matching.}
\label{fig:abstract}
\end{figure}

In this paper, we aim to interpret the success of deepfake detection models from the novel perspective of image matching. We consider the matching images as follows. 
As shown in Fig \ref{fig:abstract}, the face of the source image is manipulated with representations of the target image to generate the corresponding fake image.
Then the above \textbf{f}ake image, \textbf{s}ource image and \textbf{t}arget image are considered as the matching images, termed as the FST-Matching. 
To this end, we design different metrics to quantitatively evaluate the effectiveness of image matching and propose three hypotheses as follows. 

\textbf{Hypothesis 1: Deepfake detection models indicate real/fake images based on visual concepts that are neither source-relevant nor target-relevant, that is, considering such visual concepts as artifact-relevant.}
In this paper, visual concepts represent the image regions such as the mouths, noses or eyes of human faces. 
Intuitively, fake images are generated from visual concepts that are either from source images or target images. 
However, some visual concepts may inevitably be manipulated by deepfake methods, causing them to be different from both source images and target images.
Well-trained deepfake detection models are supposed to indicate real/fake images based on both source-irrelevant and target-irrelevant visual concepts. 

\textbf{Hypothesis 2: Besides the supervision of binary labels, deepfake detection models implicitly learn artifact-relevant visual concepts thr-ough the FST-Matching in the training set.} 
Intuitively, binary labels are not sufficient enough to accomplish the deepfake detection task. 
Training images usually contain other artifact-irrelevant visual concepts, such as the identity of images. 
Such visual concepts may co-appear on certain real/fake images, causing deepfake detection models to learn biased representations of the forgeries.
For example, deepfake detection models may infer the results based on the gender of images if real images are all male and fake images are all female. 
To this end, FST-Matching images are supposed to help deepfake detection models to discard artifact-irrelevant visual concepts and focus on artifact-relevant visual concepts, since they share common artifact-irrelevant visual concepts but are annotated with opposite labels. 

\textbf{Hypothesis 3: Implicitly learned artifact visual concepts through the FST-Matching in the raw training set are vulnerable to the video compression.}
Deepfake detection models trained on raw images usually suffer from significant performance drop when testing on compressed images \cite{face-x-ray,amazon,ff++}. 
We assume that it is because the implicit learning of artifact visual concepts through FST-Matching is fragile to the video compression.
Specifically, the implicitly learned artifact visual concepts may become indistinguishable from compressed source visual concepts and target visual concepts on fake images due to the compression, causing deepfake detection models to make false predictions.

\textbf{Methods:}
To verify the proposed hypotheses, we propose an explanation method based on the Shapley value \cite{shapley-value} to interpret the predictions of deepfake detection models with various backbones. 
The Shapley value was firstly proposed in game theory \cite{shapley-value} and is widely used in recent studies \cite{lundberg2017unified,ancona2019explaining} to interpret the representations inside DNNs.
Specifically, the Shapley value unbiasedly estimates the contributions of each player to the total award of the game.
It naturally satisfies four properties, \emph{i.e.} the linearity property, the dummy property, the symmetry property, and the efficiency property \cite{shapley-properties}, which ensures its fairness and trustworthiness.
Based on the Shapley value, we evaluate the visual concepts on images from the novel perspective of image matching to verify the proposed hypotheses.

Furthermore, during the verification of hypotheses, we surprisingly find the learned source/target visual concepts are more consistent among compressed images than the implicitly learned artifact visual concepts on images.
Combined with the understanding of hypothesis 1, we then devise a simple model by disentangling source/target-irrelevant representations from the source/target visual concepts to indicate images (termed as the FST-Matching Deepfake Detection Model), which aims to boost the performance of the forgery detection on compressed videos.
Results in our experiments show that such simple architecture achieves great performance, especially on highly compressed (\emph{e.g.} c40) videos.

\textbf{Contributions:}
Our contributions can be summarized as follows.

1. We propose a method to interpret the success of deepfake detection models from the novel perspective of image matching, \emph{i.e.} the FST-Matching.

2. Three hypotheses from the perspective of the FST-Matching are proposed and verified, which offers new insights into the task of deepfake detection.

3. We further propose the FST-Matching Deepfake Detection Model to improve the performance on compressed videos.
\section{Related work}
\subsection{Deepfake detection}
The goal for deepfake detection is to classify the input media as either real or fake. 
Previous studies of deepfake detection mainly focused on improving the model performance on various datasets.
Some methods \cite{mesonet,bayar2016deep,cozzolino2017recasting,nguyen2019use,rahmouni2017distinguishing,ff++} considered it as a binary classification task and directly trained models on the largely-collected dataset, such as Celeb-DF \cite{celeb}, DFDC \cite{dfdc}, FF++ \cite{ff++} and \emph{etc}.
These methods achieved great performance on the in-dataset evaluation, \emph{i.e.} testing models on images manipulated by learned deepfake methods.
However, these methods often failed to detect unseen datasets with newly proposed deepfake methods. 
To this end, other studies \cite{zhao2021learning,multi,zhou2021joint,hu2021dynamic} aim to increase the generalization of deepfake detection models.
These methods usually assumed that fake images share common human-perceived artifact representations introduced in the process of deepfake methods, such as blending boundaries \cite{face-x-ray}, geometric features \cite{sun2021improving} and frequency features \cite{liu2021spatial,luo2021generalizing,li2021frequency,gu2021exploiting}.
However, such assumptions usually represent human's understanding of artifact representations and may not hold in all real-life scenarios.
It still presents continuous challenges to correctly understand the key differences between real and fake images, \emph{i.e.} exploring the essence of the artifact representations on images.

To the best of our knowledge, studies focused on interpreting the learned representations of deepfake detection models are rare.
In this paper, we aim to interpret deepfake detection models from the novel perspective of image matching to demonstrate what artifact representations are to deepfake detection models, how they learned artifact representations and how to further boost their performance in real-life scenarios.

\subsection{Interpretability of DNNs}
Previous studies on the interpretability of DNNs can be roughly divided into two categories.
Some studies \cite{zeiler2014visualizing,mahendran2015understanding,dosovitskiy2016inverting,simonyan2013deep,yosinski2015understanding,zhou2016learning} focused on semantic explanations for DNNs by visualizing the learned visual concepts. 
Grad-CAM \cite{grad-cam} and Grad-CAM++ \cite{chattopadhay2018grad} explored the attribution maps of input images based on gradient information.
Zhou \emph{et. al.} \cite{zhou2014object} visualized the actual receptive fields of various units inside the DNNs.
Fong \emph{et. al.} \cite{fong2018net2vec} explored the relationship between multiple filters and learned semantic visual concepts.
Zhang \emph{et. al.} proposed to explore the relationships between the learned semantic visual concepts of DNNs via a graph model\cite{zhang2018interpreting} and a decision tree \cite{Zhang_2019_CVPR}. 
However, different from general classification tasks, deepfake detection models aim to learn artifact-relevant visual concepts on images. 
Such representation is often imperceptible to people, making it difficult to evaluate the correctness of the explanation results derived from the above methods.
Moreover, other studies proposed to explain the representations of DNNs mathematically to refrain from human evaluation of semantic representation.
To this end, some studies proposed to understand DNNs based on entropy-based methods \cite{guan2019towards,cheng2020explaining}.
Some studies explored the representations of DNNs from a game-theoretical view \cite{zhang2020interpreting,zhang2021building,zhang2021interpreting}. 
However, although the above methods can be theoretically applied to various types of DNNs, it still remains a challenge to further exploit the explanation results to instruct the learning of specific tasks, such as deepfake detection.

In this paper, we aim to bridge the gap between the general explanation results and learning better deepfake detection models from the novel perspective of image matching.
To this end, we designed the FST-Matching Deepfake Detection Model based on our explanation results and further boosted the performance on compressed videos. 
\section{Algorithms}
In this section, given a well-trained deepfake detection model, we aim to interpret its prediction from the novel perspective of image matching.
To this end, three hypotheses are proposed.
To verify these hypotheses, we propose an explanation method to evaluate the contributions of visual concepts on images based on the Shapley value \cite{shapley-value}. 
Please see supplementary materials for more information about the Shapley value.
\subsection{Artifact representations for deepfake detection models}
\fbox{
  \parbox{0.95\textwidth}{
      \textbf{Hypothesis 1 :} Deepfake detection models indicate real/fake images based on visual concepts that are neither source-relevant nor target-relevant, that is, considering such visual concepts as artifact-relevant.
  }
}
\\ \\
In this section, given a well-trained deepfake detection model $v_d(\cdot)$ (also termed as the detection encoder in this paper), we aim to evaluate the learned visual concepts on input images from the perspective of image matching. 
Specifically, we aim to explore what visual concepts on input images are considered as source-relevant, target-relevant and artifact-relevant. Then, we expect to evaluate the relationship between these visual concepts to verify the hypothesis.

The core challenge is to decide fairly what visual concepts are related to the source, target and artifact representations. Specifically, we do not annotate these visual concepts on images manually since it usually represents human's understanding of artifact representations, rather than the artifact representations inside the models.
To this end, we train a source encoder $v_s(\cdot)$ and a target encoder $v_t(\cdot)$ to indicate the source/target-relevant visual concepts on images.

Intuitively, each fake image shares certain common visual concepts with its corresponding source and target image.
We believe that when the source encoder $v_s$ classifies each fake image and its corresponding source image as the same category, $v_s$ would tend to focus on source-relevant visual concepts on each fake image.
The same way goes for the target encoder $v_t$.
Specifically, we use the additional attribute labels\footnote{implemented as the identity labels of images for convenience.} of images to train $v_s$ and $v_t$ for convenience.
To train the source/target encoder $v_s$/$v_t$, each fake image is considered as the same attribute label as the corresponding source/target image.
Each real image is considered as its original attribute label.

We use the Shapley value \cite{shapley-value} to evaluate the regional contributions of visual concepts on images to the prediction of each encoder. 
To reduce the computation cost, we divide the input image into $L\times L$ grids and calculate the contribution of each grid respectively. 
Let $G=\{g_{11},g_{12},..., g_{LL}\}$ denote the set of all grids. 
$\phi_{v_d}\in R^{L\times L}, \phi_{v_s}\in R^{L\times L}, \phi_{v_t}\in R^{L\times L}$ represent the contributions of all grids to the prediction of the detection encoder $v_d$, the source encoder $v_s$ and the target encoder $v_t$ respectively. 
In this way, $\phi_{v_d}, \phi_{v_s}$ and $\phi_{v_t}$ indicate the artifact, source and target visual concepts on images respectively. 
More specifically, given $\forall g_{ij}\in G$, it is considered to be artifact-relevant if $\phi_{v_d}(g_{ij}|G) > 0$ and artifact-irrelevant if $\phi_{v_d}(g_{ij}|G) \le 0$.
The same way goes for the source encoder $v_s$ and target encoder $v_t$.

Based on the grid-level contributions, we propose a metric to evaluate the relationship between the artifact-relevant visual concepts, source-relevant visual concepts and target-relevant visual concepts. 
According to the hypothesis, deepfake detection models are supposed to consider artifact-relevant visual concepts as neither source-relevant nor target-relevant.
Therefore, artifact-relevant visual concepts are supposed to barely have intersections with source/target-relevant visual concepts.
To this end, we firstly generate a mask $M_\tau = I(max(\phi_{v_s},\phi_{v_t})>\tau)$ to denote the most source/target-relevant visual concepts, where $I(\cdot)$ is the indicator function and $\tau$ is a certain threshold. $I(\cdot)$ returns 1 if the condition inside is valid, otherwise $I(\cdot)$ returns 0.
The metric is then designed to evaluate the intensities of the intersections between these visual concepts as follows.
\begin{equation}
    Q_{\tau} = \frac{(1 - M_{\tau}) \cdot \phi_{v_d}}{\sum_{g_{ij}\in G}{[1-M_{\tau}(g_{ij})]}} - \frac{M_{\tau}\cdot \phi_{v_d}}{\sum_{g_{ij}\in G}{M_{\tau}(g_{ij})}}
\end{equation}
where $\cdot$ denotes the inner product.
The first term measures the average intensities of the intersections between source/target-irrelevant visual concepts and artifact-relevant visual concepts. 
The second term measures the average intensities of the intersections between the source/target-relevant visual concepts and artifact-relevant visual concepts.
$Q_{\tau}>0$ represents that artifact-relevant visual concepts are more related to source/target-irrelevant visual concepts than the source/target-relevant visual concepts.
$Q_{\tau}<0$ represents that artifact-relevant visual concepts are less related to source/target-irrelevant visual concepts than the source/target-relevant visual concepts.
\subsection{Learning the artifact representations}
\fbox{
  \parbox{0.95\textwidth}{
      \textbf{Hypothesis 2:} Besides the supervision of binary labels, deepfake detection models implicitly learn artifact-relevant visual concepts through the FST-Matching in the training set.
  }
}
\\ \\ 
In this section, to verify the hypothesis, we expect to evaluate how the FST-Matching in the training set affects the learning of deepfake detection models. 
Specifically, FST-Matching in the training set means that real images contain the corresponding source and target images of fake images.
To this end, we train two models with the paired training set and the unpaired training set separately. 
In the paired training set, the real images are only the corresponding source images and target images of fake images. 
In the unpaired images, the real images are of the same number as real images in the paired training set but do not correspond to any fake images.
Then we compare the ACC, video-level AUC and the proposed metric $Q_\tau$ on these two models to evaluate the effectiveness of the FST-Matching.

\subsection{Vulnerability of artifact representations to video compression}
\fbox{
  \parbox{0.95\textwidth}{
      \textbf{Hypothesis 3:} Implicitly learned artifact visual concepts through the FST-Matching in the raw training set are vulnerable to the video compression.
  }
}
\\ \\
In this section, to verify the hypothesis, we aim to measure the stability of implicitly learned artifact visual concepts to the video compression.
Note that the detection encoder $v_d$ is firstly trained on raw images and tested on compressed images afterwards.
To this end, we design the stability metric to evaluate the changes among artifact visual concepts under the conditions of different compression rates \emph{i.e.} c23, c40.
The stability metric is designed as follows.
\begin{equation}
\delta_{v_d} = E_{cmp\in\{c23,c40\}}[cos(\phi_{v_d}^{cmp}, \phi_{v_d}^{raw})]
\end{equation}
where $\phi_{v_d}^{cmp}$ represents the grids contributions to the predictions of the detection encoder $v_d$ when tested on the compressed images. 
$\phi_{v_d}^{raw}$ represents the grids contributions tested on the raw images. 
$cos(\cdot, \cdot)$ denotes the operation of calculating the cosine similarity.
A smaller value of $\delta_{v_d}\in [-1,1]$ indicates that the implicitly learned artifact visual concepts are vulnerable to the compression.
Moreover, we also evaluate the stability of the learned source/target visual concepts for source/target encoder $v_s/v_t$ on compressed videos for more comparisons.

\subsection{FST-Matching Deepfake Detection Model}
\begin{figure}[t]
\centering
\includegraphics[width=0.8\textwidth]{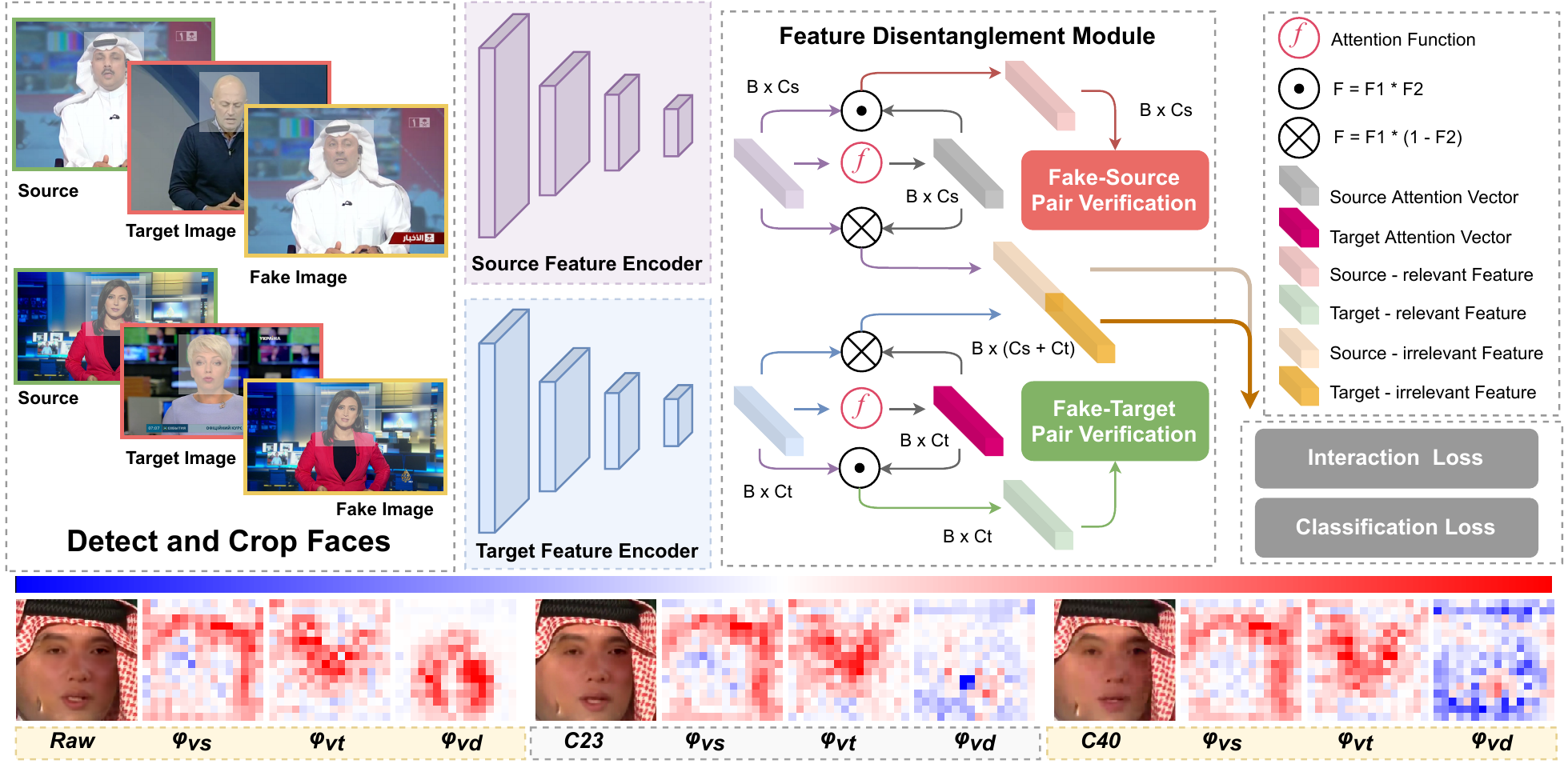}
\caption{\textbf{The FST-Matching Deepfake Detection Model.} As shown in the bottom of the figure, we surprisingly find that $\phi_{v_s}$ and $\phi_{v_t}$ are more robust to video compression than $\phi_{v_d}$. 
To this end, we use a Source Feature Encoder and a Target Feature Encoder to explicitly learn the source and target representations on images. The Feature Disentanglement Module further extracts source/target-irrelevant representations to indicate the realism of images \emph{i.e.} real or fake.}
\label{fig:FSTNet}
\end{figure}

Based on the understanding of deepfake detection models from the perspective of FST-Matching, we propose the FST-Matching Deepfake Detection Model to further boost the performance of deepfake detection models on compressed videos.
During the verification of hypothesis 3, we surprisingly found that source/target visual concepts learned by the source encoder $v_s$ and the target encoder $v_t$ (\emph{i.e.} $\phi_{v_s}$ and $\phi_{v_t}$) are more consistent than the artifact visual concepts implicitly learned by the detection encoder $v_d$ (\emph{i.e.} $\phi_{v_d}$) on compressed images (shown in the bottom of Fig \ref{fig:FSTNet}).
Inspired by the understanding of hypothesis 1, we believe that directly disentangling source/target-irrelevant representations from source/target visual concepts to indicate images may improve the model performance on compressed videos.
Please see supplementary for detailed verification.

The structure of the FST-Matching Deepfake Detection Model is shown in Fig. \ref{fig:FSTNet}, which aims to classify face forgeries based on source/target-irrelevant visual concepts on images according to hypothesis 1. 
To this end, we first use the Source Feature Encoder and the Target Feature Encoder to directly learn the source feature $f_s \in R^{B\times C_s}$ and the target feature $f_t \in R^{B\times C_t}$ on images.
$B$ indicates the number of input images.
$C_s$ and $C_t$ indicate the number of output channels.
Then we design the Feature Disentanglement Module to automatically disentangle the source/target-irrelevant feature $f_s^{ir}, f_t^{ir}$ and source/target-relevant feature $f_s^r, f_t^r$ on the channel-level.
Similar to \cite{SENet}, we use the channel-wise attention vectors $a_s \in R^{B\times C_s}$ and $a_t \in R^{B\times C_t}$ to disentangle $f_s$ and $f_t$, which are calculated as follows.
\begin{align}
    a_s = \sigma(MLP(f_s)), \ a_t = \sigma(MLP(f_t))
\end{align}
where $MLP$ denotes the multi-layer perceptron and $\sigma$ denotes the sigmoid function.
In this way, the source and target relevant feature $f_s^r, f_t^r$ are calculated as $f_s^r = a_s \circ f_s$ and $f_t^r = a_t \circ f_t$.
The source and target irrelevant feature $f_s^{ir}, f_t^{ir}$ are calculated as $f_s^{ir} = (1 - a_s) \circ f_s$ and $f_t^{ir} = (1 - a_t) \circ f_t$. Here $\circ$ denotes the channel-wise product. 

To ensure the effectiveness of the feature disentanglement, we use the Fake-Source Pair Verification module to classify $f_s^r$ as the same attribute label of the source images\footnote{implemented as the identity labels of images for convenience.}.
Similarly, $f_t^r$ is classified as the same attribute label of the target image through the Fake-Target Pair Verification module.
$f_s^{ir}$ and $f_t^{ir}$ are then concatenated to predict the final real/fake label of the input image.
Let $y_s, y_t, y_d$ denote the source attribute label, target attribute label and forgery detection label of the image. $\hat{y}_s, \hat{y}_t, \hat{y}_d$ denote the predicted source attribute, target attribute and forgery prediction.
The classification loss of the FST-Matching Deepfake Detection Model is designed as follows.
\begin{equation}
    Loss_{cls} = - E[y_d log \hat{y}_d] - \lambda_s E[y_s log \hat{y}_s] - \lambda_t  E[y_t log \hat{y}_t ]
\end{equation}

Moreover, inspired by \cite{zhang2020interpreting}, we design another loss to further strengthen the interaction between $f_s^{ir}$ and $f_t^{ir}$ for the final prediction. 
Let $h(\cdot)$ denote the final prediction module.
The interaction loss aims to increase the additional award caused by the coalition [$f_s^{ir}$, $f_t^{ir}$] \emph{w.r.t.} the sum of the award when $f_s^{ir}$ and $f_t^{ir}$ contribute to the final prediction individually. 
The interaction loss is designed as follows.
\begin{align}
    Loss_{interaction} = - E [h([f_s^{ir},f_t^{ir}]) - h([\mathbf{0},f_t^{ir}]) - h([f_s^{ir}, \mathbf{0}]) + h([\mathbf{0}, \mathbf{0}])]
\end{align}
where $\mathbf{0}$ represents the zero vector in the same size with $f_s^{ir}$ and $f_t^{ir}$. 
$h([\mathbf{0}, \mathbf{0}]$ represents the basic score when neither $f_s^{ir}$ nor $f_t^{ir}$ contributes to the final prediction. 
The overall loss is designed as follows.
\begin{align}
    Loss = Loss_{cls} + \lambda_{inter} Loss_{interaction}
\end{align}
\section{Experiment}
\subsection{Implementation details}
\textbf{DNNs \& Datasets:} To verify the proposed hypotheses, we conduct various experiments on different backbones. 
Specifically, we used ResNet-18/34 \cite{resnet} and EfficientNet-b3 \cite{efficientnet} as the backbones for the detection encoder $v_d$, $v_s$ and $v_t$. 
Besides, we also used the pre-trained models released in \cite{ff++} and \cite{multi}  for the detection encoder $v_d$ for more comparisons with state-of-the-art methods.

\begin{figure}[t]
\centering
\subfigure{
  \includegraphics[width=0.28\textwidth]{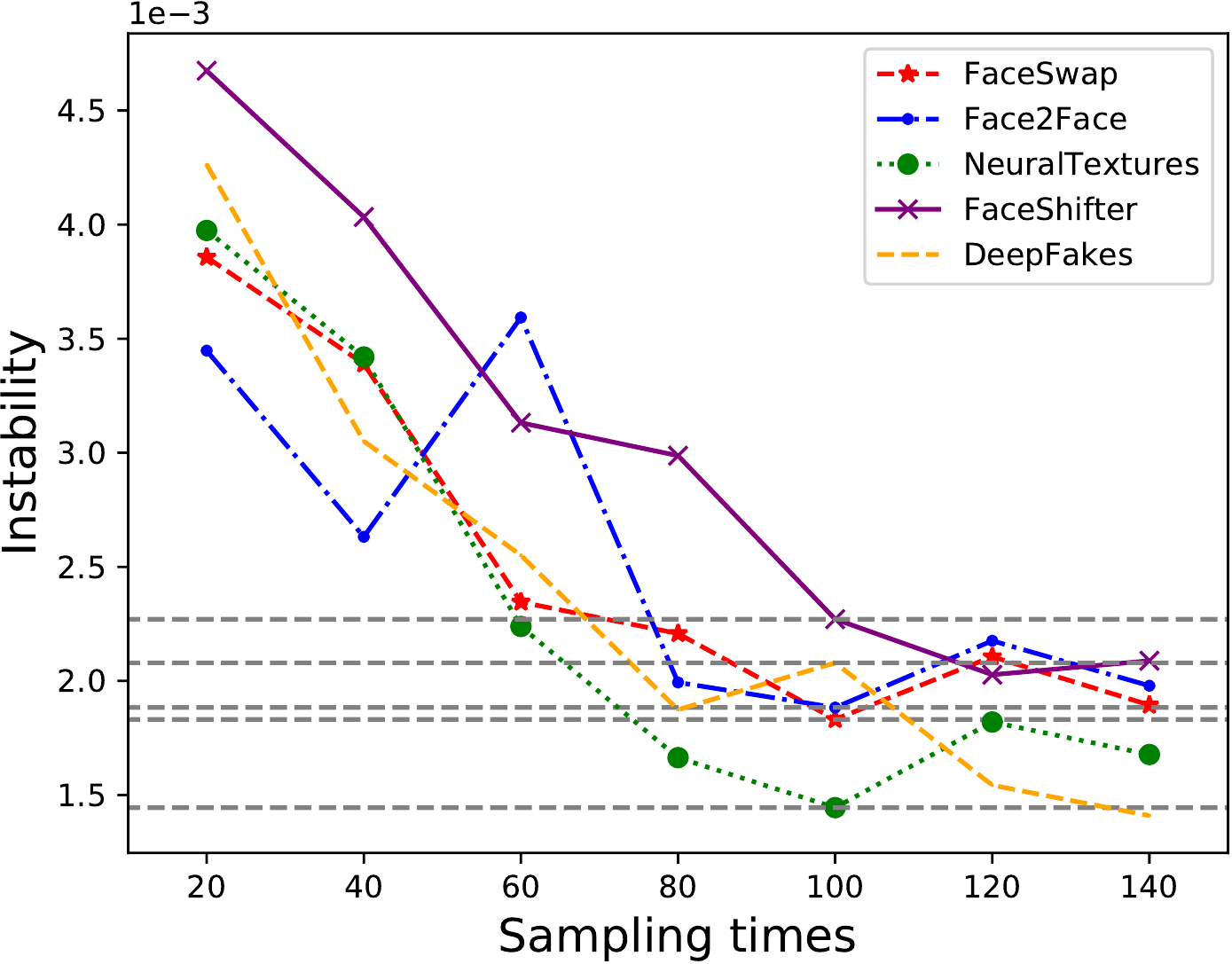}
}
\subfigure{
  \includegraphics[width=0.65\textwidth]{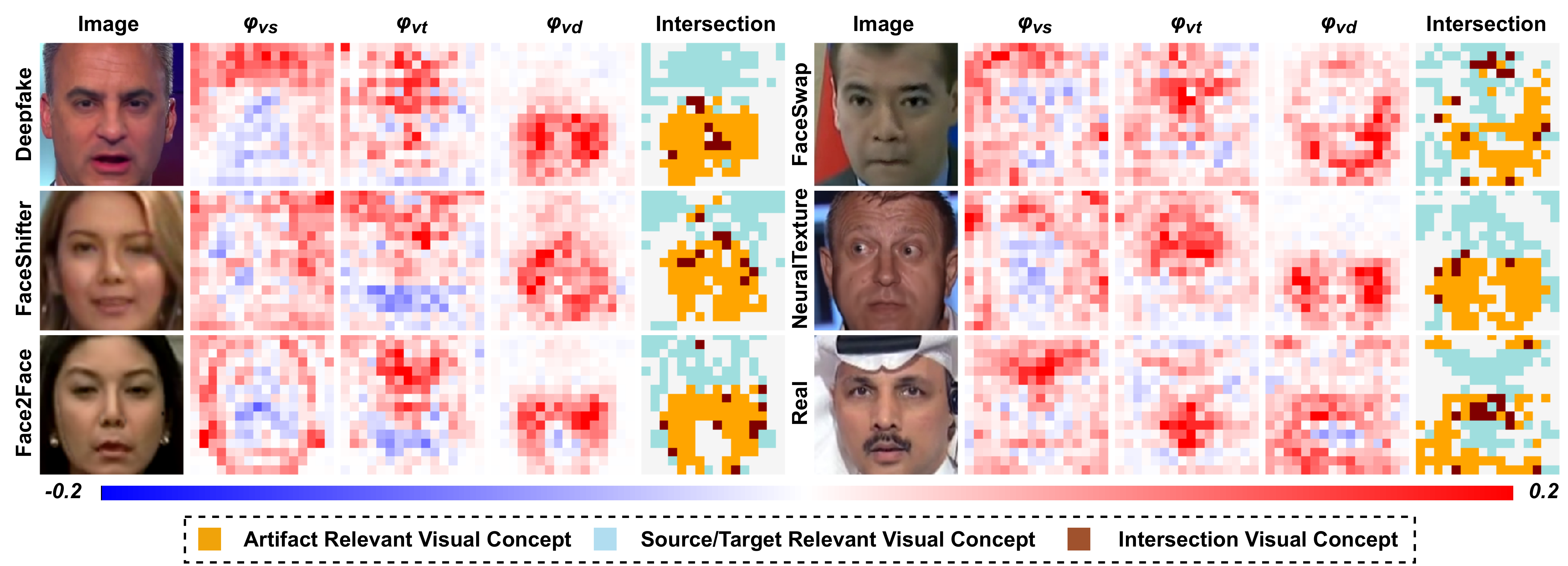}
}
\caption{\textbf{Instability of the Shapley value (left) and verification of hypothesis 1(right).}
The left figure shows that as the sampling times increase, the Shapley value becomes stable.
The right figure shows the visualization of source, target and artifact visual concepts, \emph{i.e.} $\phi_{v_s}, \phi_{v_t}$ and $\phi_{v_d}$. Results show that artifact-relevant visual concepts barely have intersections with source/target-relevant visual concepts among various manipulation algorithms, which supports hypothesis 1. }
\label{fig:HP1_vis}
\end{figure}

We trained and tested our models on the widely-used FF++ \cite{ff++} dataset. 
FF++ \cite{ff++} dataset contains  5000 videos, including 1000 original videos and 4000 fake videos manipulated by different forgery methods, such as Deepfake \cite{deepfakes}, FaceSwap \cite{faceswap}, FaceShifter \cite{faceshifter}, NeuralTextures \cite{neural-textural} and Face2Face \cite{face2face}. 
All models were pre-trained on the ImageNet \cite{imagenet} dataset and fine-tuned on FF++ \cite{ff++}. 
Moreover, the attribute label of the input image is set as the identity of the image for convenience.
Specifically, for the fake image, the source/target encoder is expected to classify the image as the identity of its corresponding source/target image.
For the real image, the source encoder and the target encoder are both expected to classify the image as its own original identity.

\noindent \textbf{Implementation of the Shapley value:} The precise calculation of the Shapley value is computationally intolerable.
To this end, we used the sampling-based method \cite{shapley-sample} to approximately calculate the contributions of all the visual concepts.
During the sampling process, the unsampled grids of images were set as the baseline value, which is set to be zero in this paper.
Moreover, we used the selected scalar before the softmax layer corresponding to the ground truth label of the image as the output score for all the encoders. 

\subsection{Fairness of the Shapley value}
\subsubsection{Accuracy of the Shapley value}
To ensure the stability of the approximated Shapley value, we evaluated the effect of sampling times $T$ \emph{w.r.t} the change of the Shapley value.
Specifically, similar to \cite{zhang2021building}, we repeated the sampling procedures \cite{shapley-sample} two times for the same sampling times $T$ to get $\phi_1$ and $\phi_2$ respectively.
Then we measured the change between $\phi_1$ and $\phi_2$ \emph{w.r.t} to the sampling times $T$ via the instability metric $\frac{||\phi_1 -\phi_2||_2}{||\phi_1 + \phi_2||_2}$ among all test images.
As shown in Fig \ref{fig:HP1_vis}, we calculated the instability metric for ResNet18-based $\phi_{v_{d}}$ for different samping times. 
Results show that when $T \ge 100$, we get the relatively stable Shapley value, which ensures the fairness of our results. 

\subsection{Verification of hypotheses}
\subsubsection{Verification of hypothesis 1.}
Hypothesis 1 assumes that well-trained deepfake detection models indicate images based on neither source-relevant nor target-relevant visual concepts, \emph{i.e.} considering them to be artifact-relevant. In this section, we both qualitatively and quantitatively verify the hypothesis.

For the qualitative analysis, we find that artifact-relevant visual concepts barely have intersections with source/target-relevant visual concepts. 
In Fig. \ref{fig:HP1_vis}, we showed the visual results of $\phi_{v_s}, \phi_{v_t}, \phi_{v_d}$ and the intersections among the main contributed visual concepts for different manipulation algorithms used in FF++ \cite{ff++}. 
For the better visualization, we normalized $\phi_{v_s}, \phi_{v_t}, \phi_{v_d}$ all to the unit vector.
The backbone of the detection decoder $v_d$ is ResNet-18 \cite{resnet}.
The source and target relevant visual concepts are denoted based on the mask $M_\tau$.
For more clarity, in the column of \emph{Intersection}, we only kept the top highest 30\% contributed grids.
Results show that deepfake detection models mainly consider artifact-relevant concepts as neither source-relevant nor target-relevant. 

For the quantitative analysis, we evaluated the proposed metric $Q$ among various DNNs and manipulation algorithms.
In Table \ref{table:hypothesis1}, we calculated the average value of $Q$ among different thresholds $\tau$ for a fair comparison.
Specifically, $\tau$ was set to different values to keep $\{0.60L^2, 0.65L^2, ...,0.85L^2, 0.9L^2, 0.95L^2\}$ grids on $M_{\tau}$ respectively.
$Q > 0$ represents that the learned artifact-relevant visual concepts are more related to source/target-irrelevant visual concepts than the source/target-relevant visual concepts.
Results show that various types of DNNs mainly consider artifact-relevant visual concepts as neither source-relevant nor target-relevant.
Moreover, such results are not essentially related to the choices on backbones of $v_s$ and $v_t$, which further verify the generality of the hypothesis.
Note that $Q<0$ for Xception \cite{ff++} when tested on images manipulated by FaceShifter \cite{faceshifter}. 
It is because that the originally released pre-trained models Xception in \cite{ff++} was never trained on forged images of FaceShifter \cite{faceshifter} before, thus unable to locate the artifact-relevant visual concepts for FaceShifter \cite{faceshifter}.


\begin{figure}[t]
\centering
\includegraphics[width=0.99\textwidth]{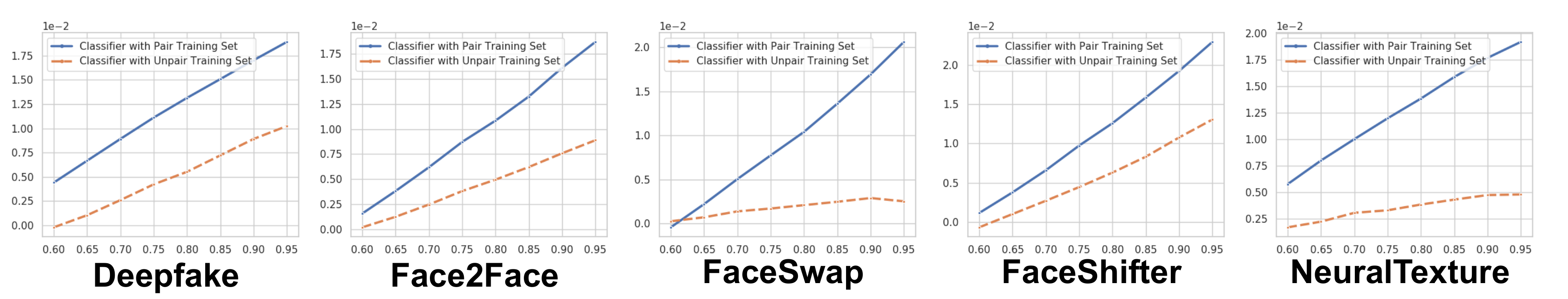}
\caption{
\textbf{Verification of hypothesis 2: }comparison of the proposed metric $Q_\tau$ between models trained on the paired training set and the unpaired train set. The horizontal coordinate represents the percentage of the kept grids in the mask $M_\tau$ when setting different thresholds $\tau$. The backbone of the detection encoder is ResNet-18 \cite{resnet}. Results show that models trained on the paired training set have larger values of $Q_\tau$, showing that FST-Matching helps models to locate artifact-relevant visual concepts.
}
\label{fig:sft_valid}
\end{figure}


\begin{table}[t]
\setlength{\tabcolsep}{3pt}
\begin{center}
\caption{\textbf{Verification of hypothesis 1:} comparison of the proposed metric $Q$ ($\times 10^{-2}$) for different deepfake detection models among various manipulation algorithms.
Results show that well-trained deepfake detection models have larger values of $Q$, which indicates that these models consider source/target-irrelevant visual concepts as artifact-relevant.}
\linespread{1.1}
\tiny
\label{table:hypothesis1}
\begin{tabular}{l | c | c  c  c  c  c }

\hline

\multirow{2}{*}{Backbone of $v_{s} / v_{t}$} 
& \multirow{2}{*}{Forgery Methods} 
& \multicolumn{5}{c}{Backbone of $v_{d}$ ($Q (\times 10^{-2})$)} \\ \cline{3-7}

~ & ~ & ResNet-18 & ResNet-34 & Efficient-b3 & MAT \cite{multi} & Xception \cite{ff++} \\ 

\hline
\hline
\multirow{5}{*}{ResNet-18 \cite{resnet}}
~ & FaceSwap \cite{faceswap}
& 2.77 & 2.88 & 2.02 & 2.57 & 3.10 \\

~ & Face2Face \cite{face2face}
& 2.31 & 2.63 & 2.08 & 2.54 & 2.59 \\

~ & FaceShifter \cite{faceshifter}
& 2.45 & 3.22 & 2.10 & 2.42 & -0.73 \\

~ & Deepfake \cite{deepfakes}
& 2.53 & 2.67 & 2.30 & 2.79 & 2.61 \\

~ & NeuralTexture \cite{neural-textural}
& 2.30 & 2.67 & 2.07 & 2.51 & 1.00 \\

\hline
\multirow{5}{*}{Efficient-b3 \cite{efficientnet}}
~ & FaceSwap \cite{faceswap}
& 2.85 & 2.99 & 2.08 & 2.49 & 3.20 \\

~ & Face2Face \cite{face2face}
& 2.19 & 2.63 & 2.00 & 2.49 & 2.61 \\

~ & FaceShifter \cite{faceshifter}
& 2.38 & 3.22 & 2.07 & 2.33 & -0.67 \\

~ & Deepfake \cite{deepfakes}
& 2.51 & 2.71 & 2.17 & 2.77 & 2.64 \\

~ & NeuralTexture \cite{neural-textural}
& 2.32 & 2.69 & 2.05 & 2.47 & 1.06 \\

\hline
\end{tabular}
\end{center}
\end{table}

\begin{table}[t]
\begin{center}
\setlength{\tabcolsep}{8pt}

\caption{\textbf{Verification of hypothesis 2:} performance comparison between models trained on the whole FF++ \cite{ff++} dataset (denoted as the \emph{Baseline}), the paired training set and the unpaired training set. In the paired training set, real images are the corresponding source and target images of fake images \emph{i.e.} satisfying the FST-Matching. 
Results show that models trained on the paired training set achieve similar performance to the baseline. 
Note that paired training set is of a significantly smaller size. 
Such results demonstrate the effectiveness of the FST-Matching.}
\linespread{1.1}
\tiny
\label{table:sft_matching}
\begin{tabular}{l | c |c c| c c| c c }
\hline
\multirow{2}{*}{Models} & \multirow{2}{*}{Forgery Methods} & \multicolumn{2}{c|}{Baseline} & \multicolumn{2}{c|}{Pair} & \multicolumn{2}{c}{Unpair} \\ \cline{3-8}

~ & ~ & \textit{ACC} & \textit{AUC} & \textit{ACC} & \textit{AUC} & \textit{ACC} & \textit{AUC}\\

\hline
\hline

\multirow{5}{*}{ResNet-18 \cite{resnet}} 
~ & FaceSwap \cite{faceswap}& 98.93 & 100 & 97.50 & 99.91 & 53.93 & 75.41 \\

~ & Face2Face \cite{face2face}& 96.79 & 99.43 & 97.14 & 99.27 & 64.29 & 85.74 \\

~ & FaceShifter \cite{faceshifter}& 99.29 & 99.99 & 97.14 & 99.82 & 81.07 & 93.03 \\

~ & Deepfake \cite{deepfakes}& 98.21 & 100 & 97.50 & 99.87 & 69.64 & 86.51 \\

~ & NeuralTexture \cite{neural-textural}& 90.71 &	98.89 & 95.71 & 98.73 & 60.00 & 76.60 \\

\hline

\multirow{5}{*}{Efficient-b3 \cite{efficientnet}} 
~ & FaceSwap \cite{faceswap}& 100 & 100 & 99.64 & 100 & 77.50 & 87.51 \\

~ & Face2Face \cite{face2face}& 99.29 & 99.77 & 99.29 & 99.72 & 81.79 & 93.36 \\

~ & FaceShifter \cite{faceshifter}& 99.29 & 99.93 & 99.29 & 99.96 & 84.29 & 96.10 \\

~ & Deepfake \cite{deepfakes}& 100 & 100 & 100 & 100 & 85.36 &  97.81 \\

~ & NeuralTexture \cite{neural-textural}& 99.29 & 99.85 & 98.93 & 99.56 & 82.86 & 92.30 \\
\hline
\end{tabular}
\end{center}
\end{table}
\subsubsection{Verification of hypothesis 2.}
Hypothesis 2 assumes that well-trained deepfake detection models implicitly learned artifact-relevant visual concepts through the FST-Matching in the training set. 
To verify the hypothesis, we trained two models of the same backbone on the paired training set and the unpaired training set separately.
In the paired training set, real images are only the source and target images corresponding to the fake images. 
In contrast, the real images in the unpaired training set do not match fake images, but are of the same number as the real images in the paired training set.
\textbf{Both the paired and unpaired training set are downsampled from FF++ \cite{ff++} dataset containing only 40 identities of images, which is significantly small compared to the initial 1000 identities in the FF++ \cite{ff++} dataset.}
In this section, we conduct extensive experiments to demonstrate that FST-Matching is crucial for learning deepfake detection models.

Firstly, we compared the ACC and video-level AUC on each trained model.
As shown in Table \ref{table:sft_matching}, models trained on the paired training set achieved similar performance to the baseline models, which are trained on the whole FF++ \cite{ff++} dataset. 
Note that the paired training set is significantly smaller than the original FF++ \cite{ff++}  dataset, which demonstrates the importance of FST-Matching in the training set.
In contrast, models trained on the unpaired training set, although of the same size as the paired training set, showed apparently worse results.
Such results also show that FST-Matching in the training set is of great value to learning deepfake detection models.

Moreover, we compared the proposed metric $Q_\tau$ between each trained model as well. 
To make a fair comparison, we calculated the value of the metric $Q_\tau$ of different $\tau$ among all the test images.
As shown in Fig \ref{fig:sft_valid}, models trained on the paired training set have larger values of $Q_\tau$, showing that FST-Matching in the training set effectively helps models to localize source/target-irrelevant visual concepts and consider them as artifact-relevant.

\begin{table}[t]
\setlength{\tabcolsep}{5pt}

\begin{center}
\caption{\textbf{Verification of hypothesis 3:} comparisons between the stability metric $\delta$ of different visual concepts. 
The backbones of the source,target and detection encoders are all ResNet-18 \cite{resnet}. 
Results show that learned source and target visual concepts are more consistent to video compression than implicitly learned artifact visual concepts.}
\linespread{1.0}
\scriptsize
\label{table:hypothesis3}
\begin{tabular}{l | c  c  c  c  c }

\hline

\multirow{2}{*}{Visual Concept} & \multicolumn{5}{c}{Forgery Methods ($\delta$)} \\ \cline{2-6}
~ & FaceSwap & Face2Face & FaceShifter & Deepfake & NeuralTexture \\ 

\hline
\hline

Source
& 0.73 & 0.74 & 0.73 & 0.74 & 0.74 \\

Target
& 0.73 & 0.76 & 0.71 & 0.75 & 0.76 \\

\hline
\hline

Artifact (Baseline)
& 0.17 & -0.02 & 0.14 & -0.15 & -0.14 \\

\hline
\end{tabular}
\end{center}
\end{table}


\subsubsection{Verification of hypothesis 3.}
Hypothesis 3 assumes that the implicitly learned artifact visual concepts through the FST-Matching in the raw training set are vulnerable to the video compression. 
To verify the hypothesis, we tested the raw-trained models on compressed videos and calculated the proposed metric $\delta_{v_d}$ among all test images.
For the qualitative analysis, as shown in Fig \ref{fig:FSTNet}, raw-trained models indicate compressed images with significantly different visual concepts compared with the raw images.
For the quantitative analysis, in Table \ref{table:hypothesis3}, the calculated $\delta_{v_d} \in [-1, 1]$ is near $0$, which also indicates the great change of $\phi_{v_d}$ under the condition of different compression rate.

Moreover, we also evaluated the stability of the source/target visual concept.
Surprisingly, as Fig \ref{fig:FSTNet} and Table \ref{table:hypothesis3} show, such learned visual concepts show great consistency to the video compression, compared to the implicitly learned artifact visual concepts. 
Such results motivate us to improve the model performance on compressed videos by devising a model, which explicitly exploits the FST-Matching in the training set.

\begin{table}[t]
\begin{center}
\setlength{\tabcolsep}{10pt}

\caption{
Performance comparison on compressed videos with state-of-the-art methods.
Our method achieves great performance on compressed videos, especially on c40 videos.}
\linespread{1.1}
\scriptsize
\label{table:cmp_sota}
\begin{tabular}{l | c | c c| c c }
\hline
\multirow{2}{*}{Models} & \multirow{2}{*}{Backbone} &  \multicolumn{2}{c|}{C23} & \multicolumn{2}{c}{C40} \\ \cline{3-6}
~ & ~ & \textit{ACC} & \textit{AUC} & \textit{ACC} & \textit{AUC}\\

\hline
\hline

Steg.Features \cite{fridrich2012rich} & -  & 70.97 & - & 55.98 & - \\

LD-CNN \cite{cozzolino2017recasting} & -  & 78.45 & - & 58.69 & - \\

Face-x-ray \cite{face-x-ray} & HRNet 
& - & 87.30 & - & 61.60 \\

MesoNet \cite{mesonet} & Xception & 83.10 & - & 70.47 & - \\

Xception \cite{ff++} & Xception & 92.39 & 94.86 & 80.32 & 81.76 \\

Xception-ELA \cite{gunawan2017development} & Xception & 93.86 & 94.80 & 79.63 & 82.90 \\

Xception-PAFilters \cite{chen2017jpeg} & Xception & - & - & 87.16 & 90.20 \\

SPSL \cite{liu2021spatial} & Xception & 91.50 & 95.32 & 81.57 & 82.82 \\

MAT-Xception \cite{multi} & Xception  & 96.37 & 98.97 & 86.95 & 87.26 \\

MAT-Efficient \cite{multi} & Efficient-b4  & \textbf{97.60} & \textbf{99.29} & 88.69 & 90.40 \\

\hline
\hline

\multirow{4}{*}{FST-Matching (ours)}  & ResNet-18 & 94.52 & 98.34 & \textbf{88.92} & \textbf{92.02} \\
~ & Xception & 94.05 & 98.27 & 87.38 & 90.44 \\
~ & Efficient-b3 & 95.95 & 98.75 & 87.62 & 90.89 \\
~ & Efficient-b4 & 96.19 & 98.81 & 88.69 & 91.27 \\

\hline
\end{tabular}
\end{center}
\end{table}

\subsection{FST-Matching Deepfake Detection Model}
\noindent \textbf{Performance comparison on compressed videos.}
In this section, we compared the performance of our model to current state-of-the-art methods.
Table \ref{table:cmp_sota} shows the performance on compressed videos.
Specifically, when aligned with the same backbone of other methods, our model achieved great performance on compressed videos, especially on highly-compressed (\emph{e.g.} c40) videos.
Such results also indicate the broad applicability of our method.
Meanwhile, note that there still exists a slight performance gap with MAT \cite{multi} on c23 in Table \ref{table:cmp_sota}.
Different from our method, MAT \cite{multi} designed specific modules to learn the frequency features of images. 
Such features are widely shown to be effective to enhance the performance of deepfake detection models on compressed videos \cite{gu2021exploiting,li2021frequency,liu2021spatial,luo2021generalizing}. 
To this end, we believe that integrating such features into our model may potentially fill this performance gap. 
Moreover, since our method is merely the first attempt to exploit our innovative explanation results, we believe that more effective methods could be further inspired based on our study in the future.

\noindent \textbf{Performance comparison on raw videos.}
In order to have a more comprehensive analysis, we also evaluated our models on raw videos.  
Results in Table \ref{table:cmp_sota_raw} show that our method still performed well on raw images.

\noindent \begin{minipage}{\textwidth}
\begin{minipage}[t]{0.49\textwidth}
\makeatletter\def\@captype{table}
\begin{center}
\caption{Evaluation on raw videos.}

\linespread{1.0}
\scriptsize
\label{table:cmp_sota_raw}
\begin{tabular}{l | c | c c}
\hline
\multirow{2}{*}{Models} & \multirow{2}{*}{Backbone} &  \multicolumn{2}{c}{RAW} 
\\ \cline{3-4}
~ & ~ & \textit{ACC} & \textit{AUC} \\

\hline
\hline

Face-x-ray \cite{face-x-ray} & HRNet & - & 98.80  \\

MesoNet \cite{mesonet} & Xception & 95.23 & -  \\

Xception \cite{ff++} & Xception & \textbf{99.26} & 99.20  \\

Xception-ELA \cite{gunawan2017development} & Xception & 98.57 & 98.40  \\

MAT-Efficient \cite{multi} & Efficient-b4 & 97.77 & 99.61 
\\

\hline
\hline

\multirow{4}{*}{FST-Matching (ours)} & ResNet-18 & 98.14 & 99.72  \\
 & Xception & 98.71 & 99.91  \\
 &  Efficient-b3  & 98.93 & 99.90  \\
 &  Efficient-b4  & 99.00 & \textbf{99.92}  \\
\hline
\end{tabular}
\end{center}
\end{minipage}
\begin{minipage}[t]{0.49\textwidth}
\makeatletter\def\@captype{table}
\begin{center}
\caption{Cross-dataset evaluation.
}

{
\linespread{1.21}
\scriptsize

\begin{tabular}{l | c |c }

\hline

Models & Backbones & Celeb-DF \\

\hline
\hline

Xception \cite{ff++} & Xception &49.03\\

SPSL \cite{liu2021spatial} & Xception & 76.88 \\

MAT \cite{multi} & Efficient-b4 & 68.44 \\

Face-x-ray \cite{face-x-ray} & HRNet & 80.58  \\

\hline
\hline

\multirow{4}{*}{FST-Matching (ours)} & ResNet-18 
& 86.00 \\
& Xception
& 88.44 \\
& Efficient-b3 
&\textbf{89.39} \\
& Efficient-b4 & 88.13 \\
\hline

\end{tabular}

\label{table:cmp_sota_cross}

}
\end{center}
\end{minipage}
\end{minipage}
\begin{table}[h]
\begin{center}
\caption{Robustness evaluation to image editing in terms of AUC ($\%$) on FF++. }
{
\linespread{1.0}
\setlength\tabcolsep{6 pt}
\scriptsize
\begin{tabular}{l |c c c c c c c c c}

\hline

Method & Saturation & Contrast & Block & Noise & Blur & Pixel & \textbf{Avg} \\

\hline
\hline

Xception \cite{ff++} & 99.3&98.6&99.7&53.8&60.2&74.2&81.0\\

Face-x-ray \cite{face-x-ray} & 97.6&88.5&99.1&49.8&63.8& 88.6& 81.2\\

LipForensices \cite{haliassos2021lips} & \textbf{99.9}&99.6&87.4&73.8&96.1& 95.6& 92.1\\
\hline
\hline

FST-Matching (ours) & 99.6&\textbf{99.9}&\textbf{99.9}&\textbf{84.8}&\textbf{99.2}&\textbf{98.7} &\textbf{97.0}\\
\hline

\end{tabular}
\label{tb:compare-sota-editing}

}
\end{center}
\end{table}

\noindent \textbf{Evaluation on the generalization ability.}
We conduct another experiment to evaluate the generalization ability of our method. 
To this end, we followed the same cross-dataset experimental setting in SPSL \cite{liu2021spatial}.
Results are shown in Table \ref{table:cmp_sota_cross}, where the metric is AUC ($\%$). 
Our models trained on FF++ \cite{ff++} achieved great performance on Celeb-DF \cite{celeb}, regardless of different backbones. \cite{liu2021spatial}. 

\noindent \textbf{Robustness to image editing operations.}
We conduct another experiment to evaluate our method when image editing operations are applied to images. 
To this end,  we followed the same robustness experiment setting in LipForensics \cite {haliassos2021lips}. 
Results are shown in Table \ref{tb:compare-sota-editing}, where the metric is AUC ($\%$). 
Our method also demonstrated great robustness to listed perturbations.
\section{Conclusions}
In this paper, we interpret the success of deepfake detection models from the novel perspective of image matching.
To this end, three hypotheses are proposed and verified among various DNNs, \emph{i.e.} 1. Deepfake detection models indicate real/fake images based on visual concepts that are neither source-relevant nor target-relevant, that is, considering such visual concepts as artifact-relevant.\
2. Besides the supervision of binary labels, deepfake detection models implicitly learn artifact-relevant visual concepts through the FST-Matching in the training set.\
3. Implicitly learned artifact visual concepts through the FST-Matching in the raw training set are vulnerable to video compression. \
Based on the understanding, we further propose the FST-Matching Deepfake Detection Model and achieve great performance on the compressed videos. This research provides an opportunity to explore the essence of artifact representation of images and sheds new light on the task of deepfake detection.

%
%

\renewcommand\thesection{\Alph{section}}
\setcounter{figure}{4}
\setcounter{table}{7}
\setcounter{section}{0}
\section{Preliminaries: the Shapley value}
Originally introduced in game theory \cite{shapley-value}, the Shapley value was used to distribute the total \emph{award/contribution} obtained by all players to each individual fairly.
Specifically, given the set of $n$ input players $N=\{1,2,...,n\}$ who participate in the game $v$, they can obtain the \emph{score} $v(N)$.
Here, the game $v$ is formulated as a function to map any participating players to a real number.
The \emph{award} obtained by players $N$ is then calculated as $v(N)-v(\emptyset)$, where
$v(\emptyset)$ is considered as the baseline \emph{score} when no players participate in the game $v$.
In order to fairly allocate the overall \emph{award}, the Shapley value $\phi(i|N)$ is calculated as the average marginal \emph{award} obtained by player $i$, when player $i$ joined any potential subset $S \subseteq {N\backslash\{i\}}$, \emph{i.e.} $v(S\cup \{i\}) - v(S)$.
In this way, the Shapley value  $\phi(i|N)$ is calculated as follows.
\begin{equation}
    \phi_v(i|N) = \sum_{S\subseteq N\backslash\{i\}}\frac{\left|S\right|! \left|N-1-S\right|!}{|N|!}(v(S\cup \{i\})-v(S))
    \label{shapley}
\end{equation}

Moreover, the Shapely value satisfies four properties to ensure its fairness and trustworthiness \cite{shapley-properties}:
\begin{itemize}
\item \emph{Linearity property:} Considering three games $u, v$ and $w$, where $u, v$ are combined as $w$. If such games satisfy $w(S)=u(S) + v(S)$, then the Shapley value of each player $i$ in the game $w$ can be combined by the Shapley value of each player $i$ in the game $u$ and the game $v$, \emph{i.e.} $\phi_w(i|N) = \phi_u(i|N) + \phi_v(i|N)$.
\item \emph{Dummy property:} If $v(S\cup \{i\}) - v(S) = 0$ for any subset $S \subseteq N\backslash \{i\}$, then the player $i$ is considered as a dummy player. Its \emph{contribution} is measured as $\phi_{v}(i|N) = v(\{i\})-v(\emptyset)$, which indicates that player $i$ participates the game $v$ independently.
\item \emph{Symmetry property:} If $v({S\cup\{i\}}) = v({S\cup\{j\}})$ for any subset $S\subseteq N\backslash\{i,j\}$, then the player $i$ and player $j$ are considered to have the same \emph{contribution}, \emph{i.e.} $\phi_v(i|N)=\phi_v(j|N)$.
\item \emph{Efficiency property:} The overall \emph{award/contribution} can be added up by the \emph{award/contribution} of each player $i$, \emph{i.e.} $\sum_i \phi_v(i|N)=v(N)-v(\emptyset)$.
\end{itemize}

\section{More about verification of hypothesis 1}
In this section, we provide more results of different backbones to verify the hypothesis.
Specifically, following the same setting in Table 1, we used another model, \emph{i.e.} ResNet-34 \cite{resnet}, as the backbone of $\phi_{v_s}$ and $\phi_{v_t}$. 
Results in Table \ref{table:hypothesis1more} are consistent with Table 1, indicating that the learned artifact-relevant visual concepts of well-trained deepfake detection models are neither source-relevant nor target-relevant.
Such results further support the hypothesis.

\begin{figure}[ht]
\centering
  \subfigure[Positive correlation between $Q$ and ACC on Deepfake \cite{deepfakes}.]{
  \includegraphics[width=0.46\textwidth]{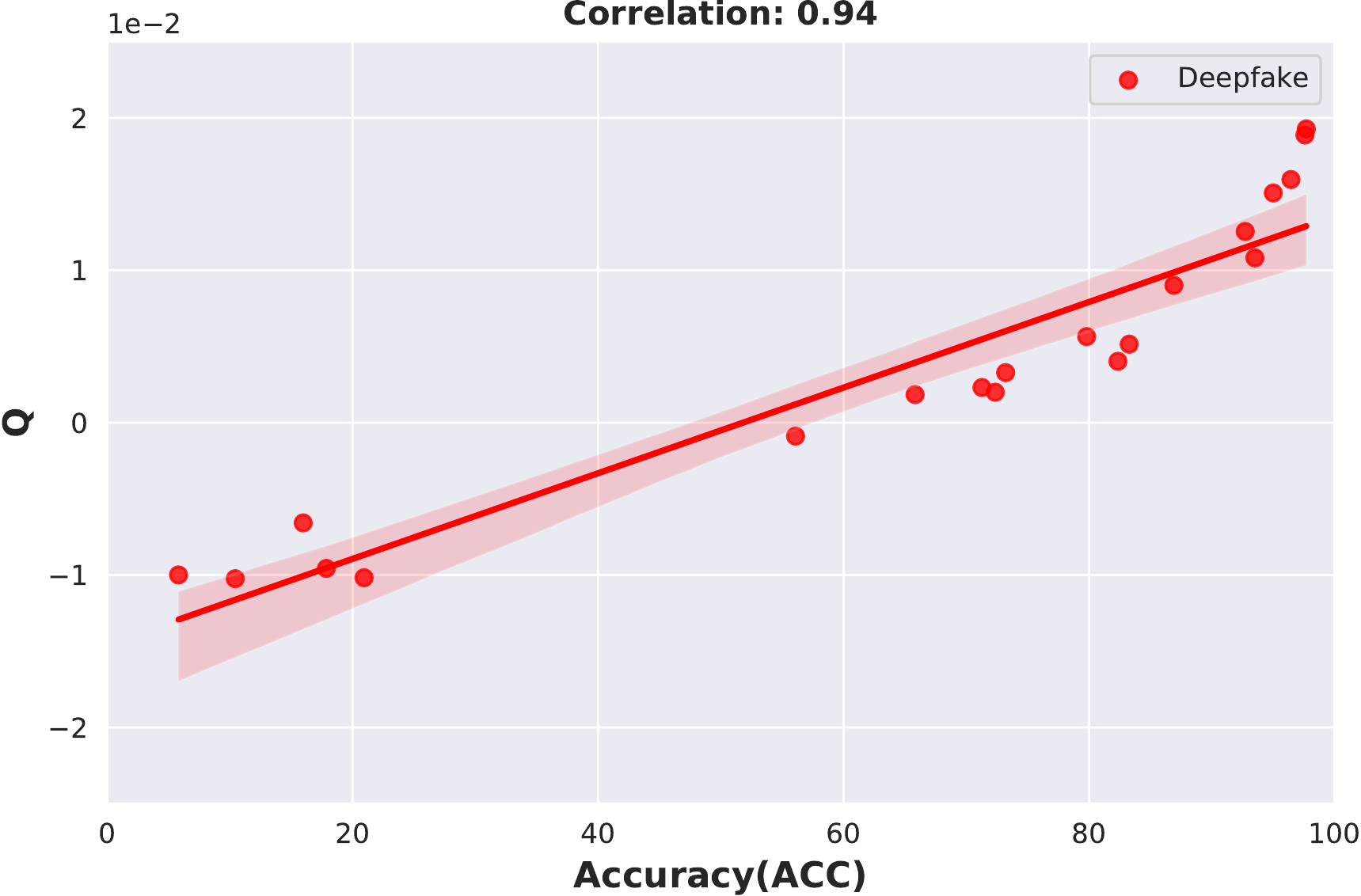}
  \label{fig:Qvis_df}
  }
  \subfigure[Positive correlation between $Q$ and ACC on Face2Face \cite{face2face}.]{
  \includegraphics[width=0.46\textwidth]{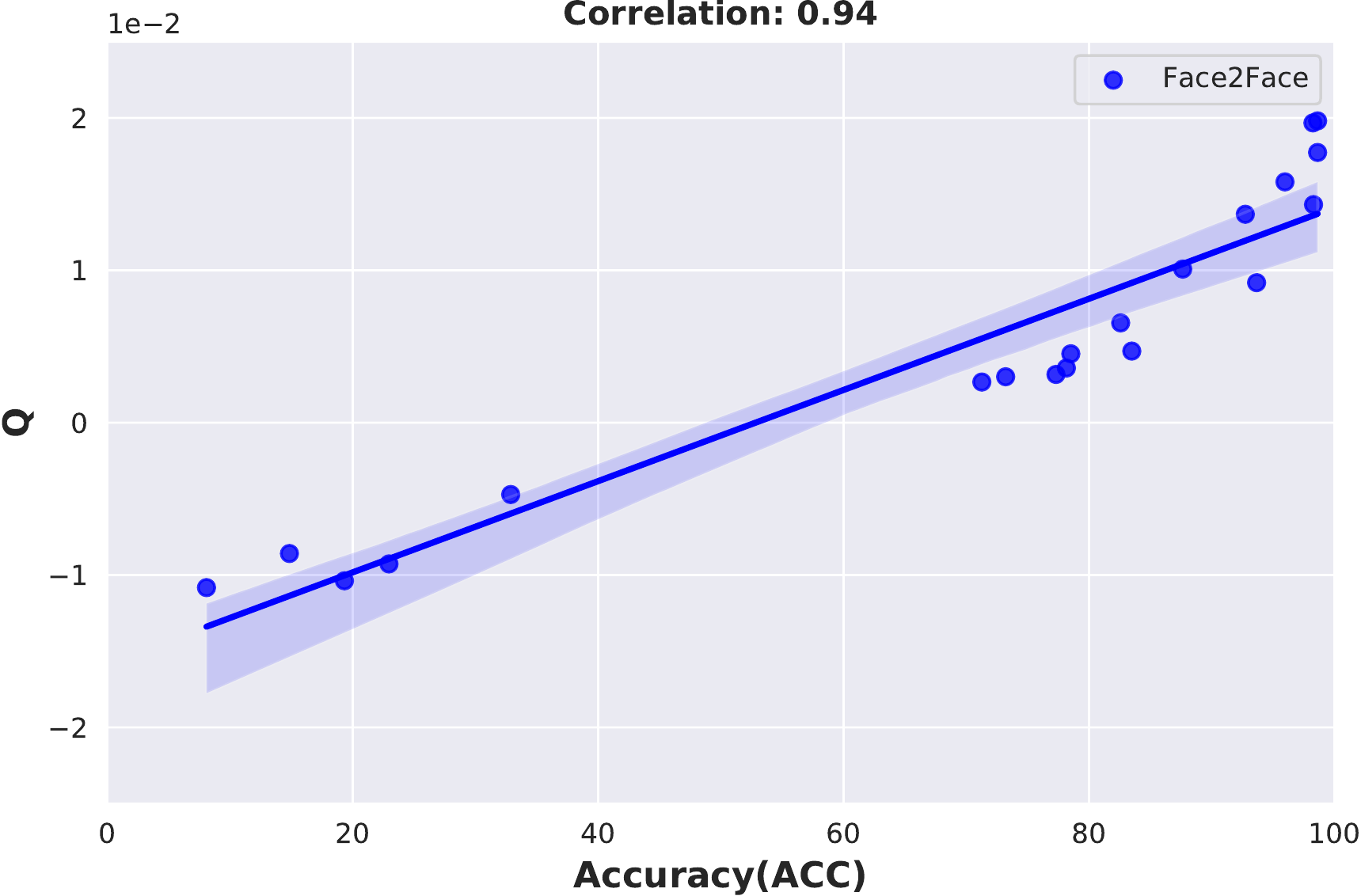}
  \label{fig:Qvis_f2f}
  }
  \newline
  \subfigure[Positive correlation between $Q$ and ACC on FaceSwap \cite{faceswap}.]{
  \includegraphics[width=0.46\textwidth]{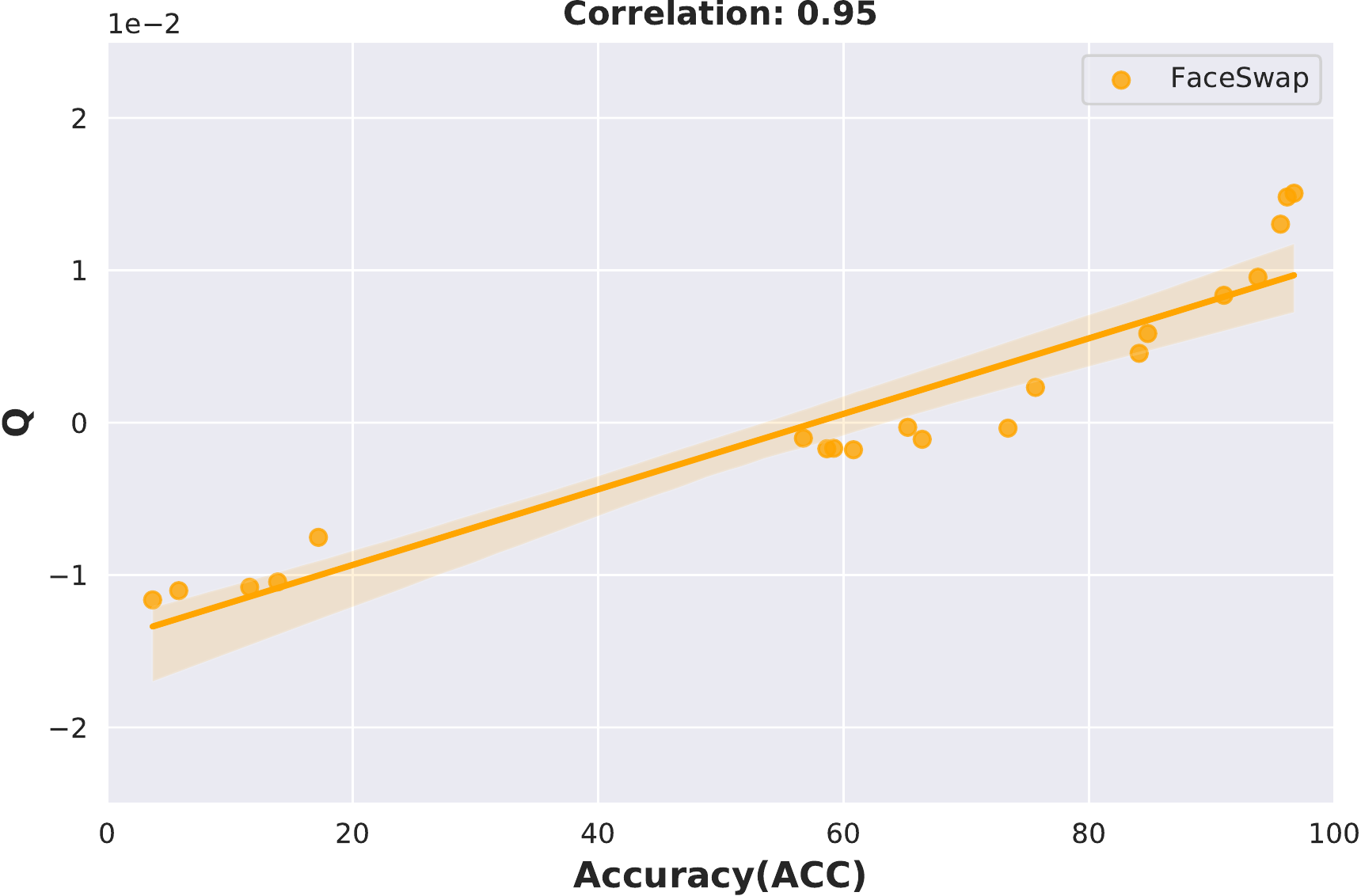}
  \label{fig:Qvis_fs}
  }
  \subfigure[Positive correlation between $Q$ and ACC on FaceShifter \cite{faceshifter}.]{
  \includegraphics[width=0.46\textwidth]{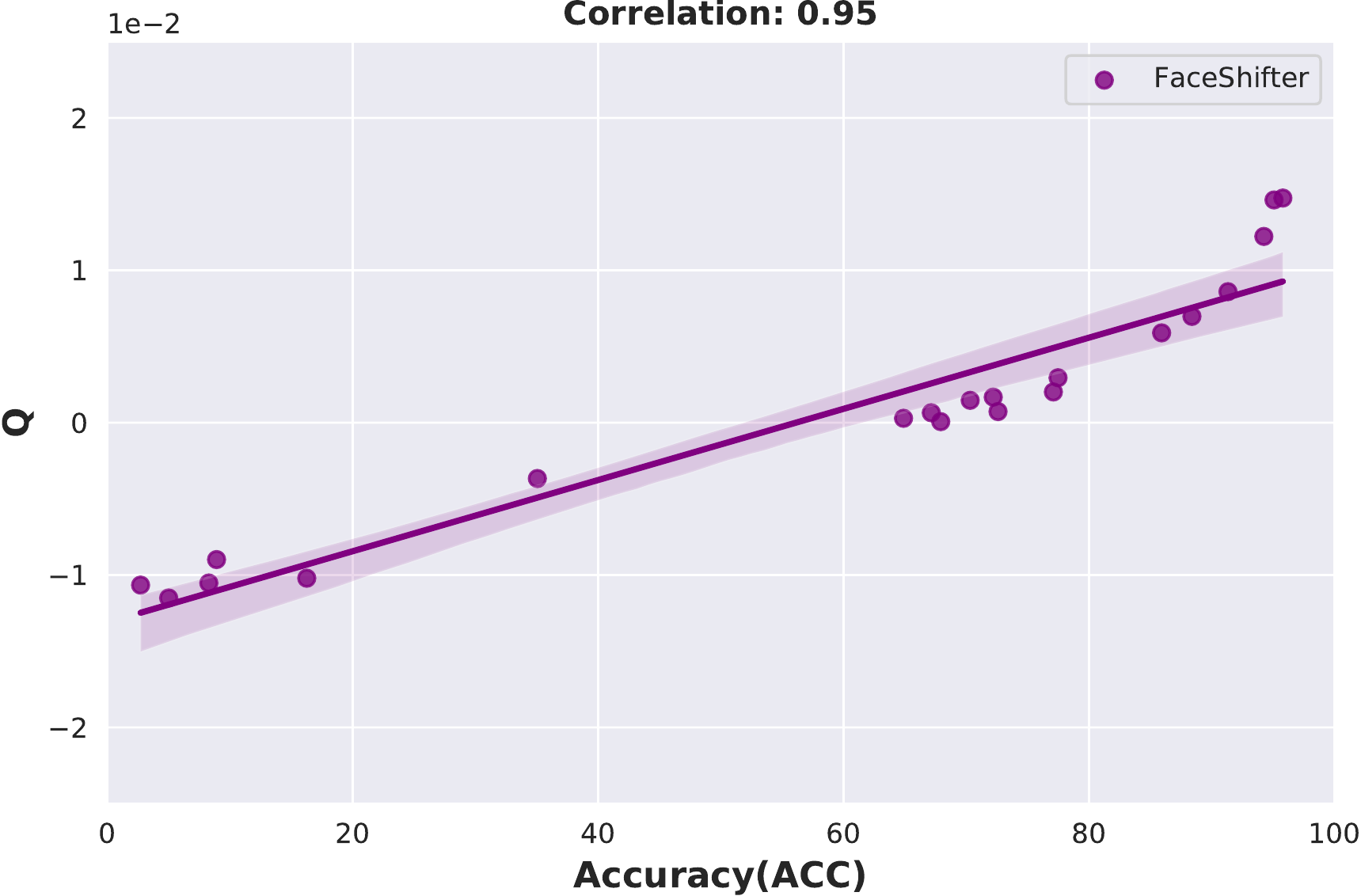}
  \label{fig:Qvis_fsh}
  }
  \newline
  \subfigure[Positive correlation between $Q$ and ACC on NeuralTexture \cite{neural-textural}.]{
  \includegraphics[width=0.46\textwidth]{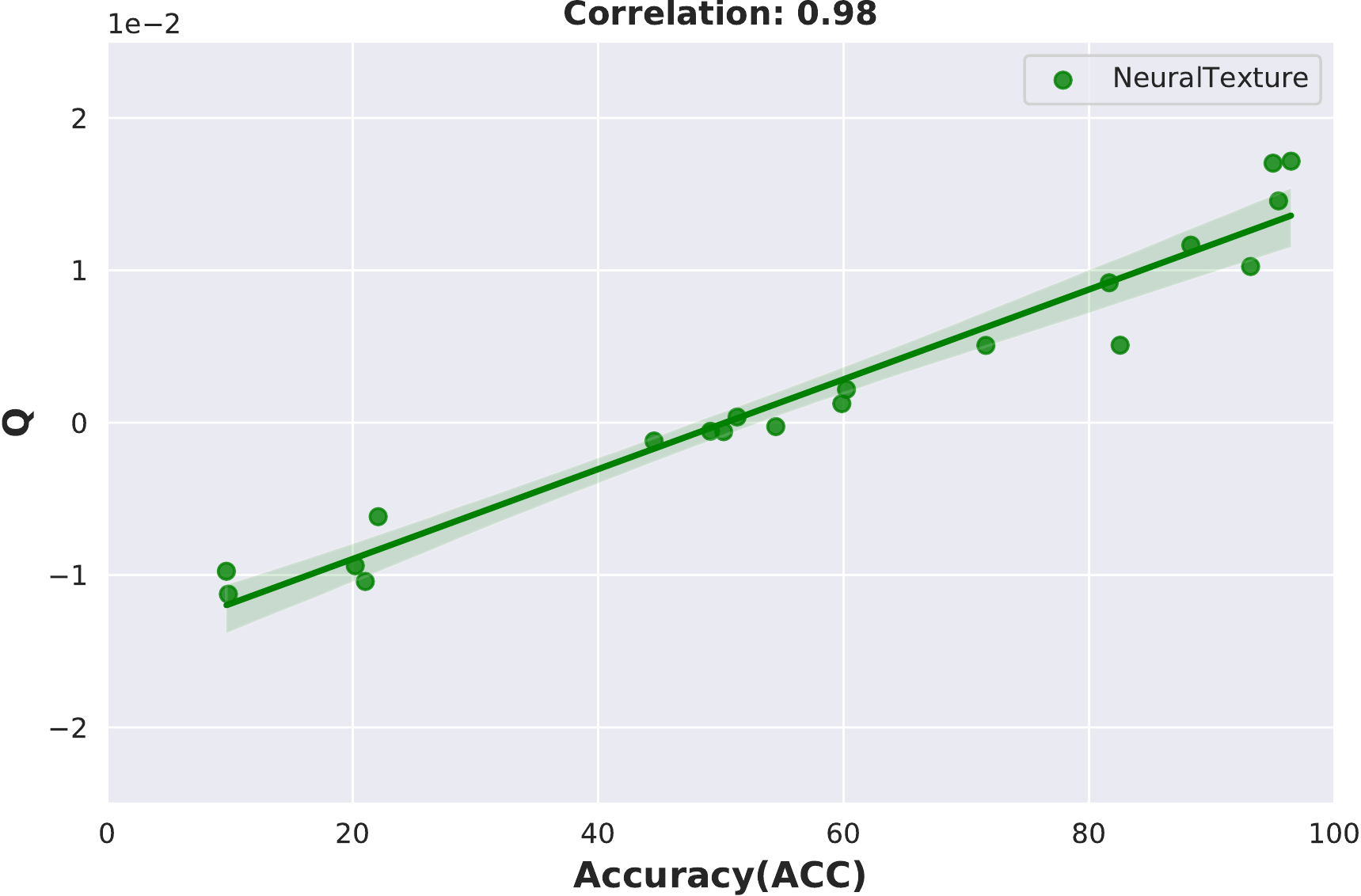}
  \label{fig:Qvis_nt}
  }
  \subfigure[Positive correlation between $Q$ and ACC on all manipulation algorithms of FF++ \cite{ff++}.]{
  \includegraphics[width=0.46\textwidth]{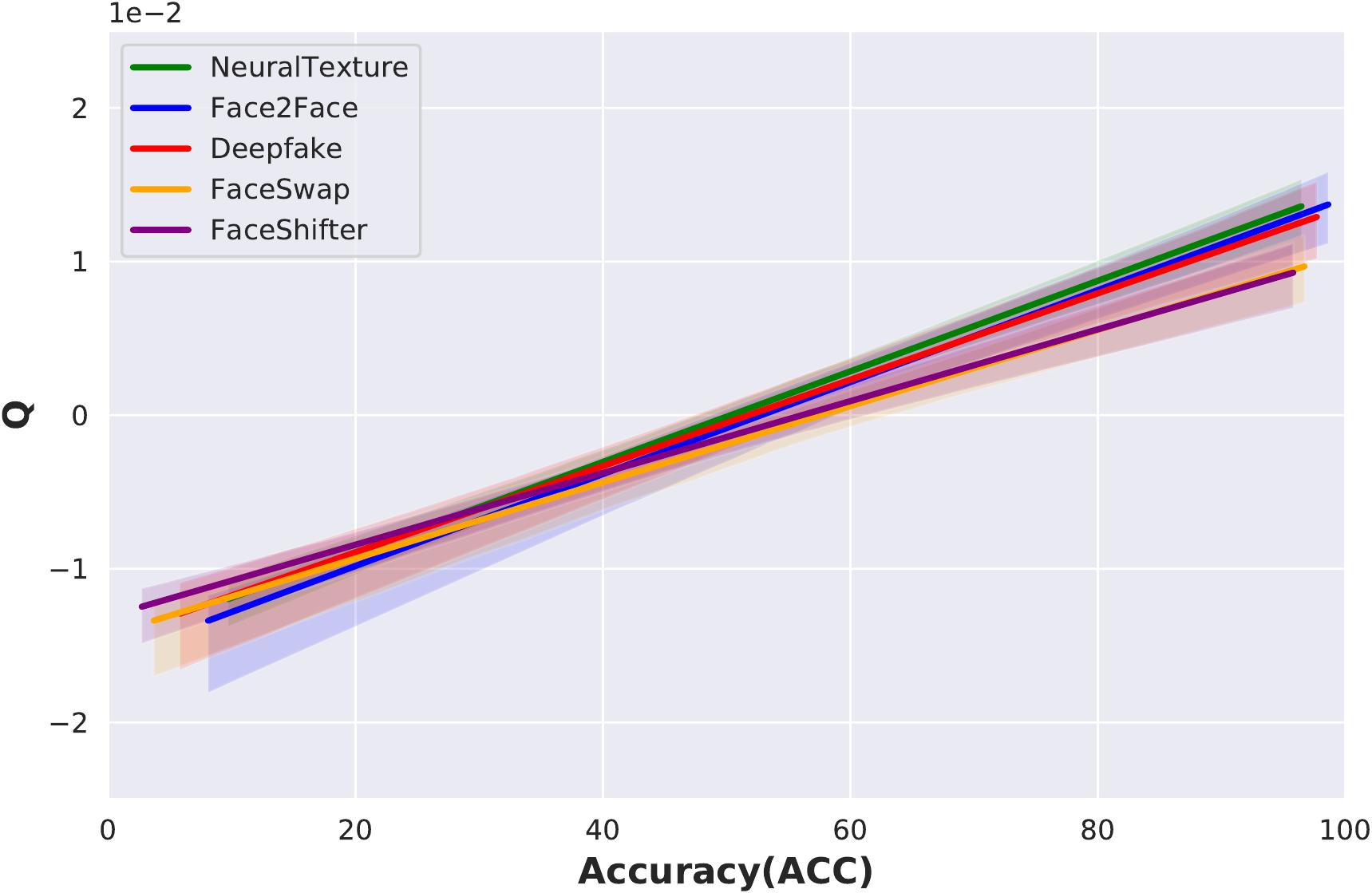}
  \label{fig:Qvis_all}
  }
  \newline
\caption{The positive correlation between the proposed metric $Q$ and the Accuracy (ACC) of deepfake detection models. 
Different points represent models of different iterations trained on FF++ \cite{ff++}. 
The correlation is calculated as the Pearson correlation. 
The backbones of models are ResNet-18 \cite{resnet}. 
Fig. \ref{fig:Qvis_all} shows that the positive correlations between the metric $Q$ and the Accuracy (ACC) are similar among different manipulation algorithms.
\textbf{Such results show that models with high accuracy consider source/target-irrelevant visual concepts as artifact-relevant.}
}
\label{fig:Qvis}
\end{figure}

Moreover, to further verify the fairness of the proposed metric $Q$, we evaluated the relationship between the proposed metric $Q$ and the Accuracy (ACC) of deepfake detection models.
Specifically, as shown in Fig. \ref{fig:Qvis}, values of $Q$ and Accuracy (ACC) of models are positively correlated.
Such results show that \textbf{when deepfake detection models achieve high accuracy, they indicate fake images based on visual concepts, which are neither source-relevant nor target-relevant. }
\begin{table}[h]
\setlength{\tabcolsep}{3pt}
\begin{center}
\caption{\textbf{More results about verification of hypothesis 1:} comparison of the proposed metric $Q$ ($\times 10^{-2}$) for different deepfake detection models among various manipulation algorithms. 
The backbones of $v_s$ and $v_t$ are all ResNet-34 \cite{resnet}.
Results are consistent with Table 1, which further supports hypothesis 1.
}
\linespread{1.6}
\tiny
\label{table:hypothesis1more}
\begin{tabular}{l | c | c  c  c  c  c }

\hline

\multirow{2}{*}{Backbone of $v_{s} / v_{t}$} 
& \multirow{2}{*}{Forgery Methods} 
& \multicolumn{5}{c}{Backbone of $v_{d}$ ($Q (\times 10^{-2})$)} \\ \cline{3-7}

~ & ~ & ResNet-18 & ResNet-34 & Efficient-b3 & MAT \cite{multi} & Xception \cite{ff++} \\ 

\hline
\hline
\multirow{5}{*}{ResNet-34 \cite{resnet}}
~ & FaceSwap \cite{faceswap}
& 2.67 & 2.80 & 2.04 & 2.53 & 2.99 \\

~ & Face2Face \cite{face2face}
& 2.07 & 2.42 & 1.96 & 2.51 & 2.40 \\

~ & FaceShifter \cite{faceshifter}
& 2.36 & 3.18 & 2.14 & 2.34 & -0.68 \\

~ & Deepfake \cite{deepfakes}
& 2.39 & 2.57 & 2.20 & 2.79 & 2.49 \\

~ & NeuralTexture \cite{neural-textural}
& 2.11 & 2.49 & 1.93 & 2.48 & 0.91 \\
\hline
\end{tabular}
\end{center}
\end{table}

\section{More about verification of hypothesis 2}
In this section, we provide more results of different backbones to verify the hypothesis.
Specifically, following the same setting in Table 2, we used ResNet-34 \cite{resnet} as the backbone and trained two models on the paired training set and unpaired training set respectively. 
Results in Table \ref{table:hypotheisi2more} are consistent with Table 2, which further support the hypothesis, indicating that the FST-Matching in the training set is of great importance to learn deepfake detection models.

\begin{table}[h]
\begin{center}
\setlength{\tabcolsep}{8pt}

\caption{\textbf{More results about verification of hypothesis 2:} performance comparison between models trained on the whole FF++ \cite{ff++} dataset (denoted as the \emph{Baseline}), the paired training set and the unpaired training set.
Results are consistent with Table 2, which further demonstrates the effectiveness of the FST-Matching.}
\linespread{1.4}
\tiny
\label{table:hypotheisi2more}
\begin{tabular}{l | c |c c| c c| c c }
\hline
\multirow{2}{*}{Models} & \multirow{2}{*}{Forgery Methods} & \multicolumn{2}{c|}{Baseline} & \multicolumn{2}{c|}{Pair} & \multicolumn{2}{c}{Unpair} \\ \cline{3-8}

~ & ~ & \textit{ACC} & \textit{AUC} & \textit{ACC} & \textit{AUC} & \textit{ACC} & \textit{AUC}\\

\hline
\hline

\multirow{5}{*}{ResNet-34 \cite{resnet}} 
~ & FaceSwap \cite{faceswap}& 98.93 & 100 & 97.14 & 99.82 & 68.21 & 71.76 \\

~ & Face2Face \cite{face2face}& 98.21 & 99.26 & 97.50 & 99.28 & 71.07 & 78.56 \\

~ & FaceShifter \cite{faceshifter}& 98.21 & 99.77 & 97.50 & 99.93 & 79.29 & 87.21 \\

~ & Deepfake \cite{deepfakes}& 98.93 & 100 & 99.29 & 99.99 & 77.14 & 82.83 \\

~ & NeuralTexture \cite{neural-textural}& 98.21 & 99.28 & 96.79 & 98.26 & 70.00 & 77.61 \\
\hline
\end{tabular}
\end{center}
\end{table}

\section{More about verification of hypothesis 3}
In this section, we provide more results of different backbones to verify the hypothesis.
Specifically, we followed the same setting in Table 3 and used ResNet-34 \cite{resnet},  EfficientNet-b3 \cite{efficientnet} as backbones.
Results in Table \ref{table:hypothesis3more} are consistent with Table 3, indicating that the learned source/target visual concepts are more robust to video compression among different backbones, compared to the implicitly learned artifact visual concepts.
Such results further support the hypothesis.
\begin{table}[h]
\setlength{\tabcolsep}{2pt}

\begin{center}
\caption{\textbf{More results about verification of hypothesis 3:} comparisons between the stability metric $\delta$ of different visual concepts. 
Results are consistent with Table 3 among different backbones, \emph{i.e.} learned source and target visual concepts are more consistent to video compression than implicitly learned artifact visual concepts.}
\linespread{1.3}
\scriptsize
\label{table:hypothesis3more}
\begin{tabular}{l | l | c  c  c  c  c }

\hline

\multirow{2}{*}{Visual Concept} & \multirow{2}{*}{Backbones} & \multicolumn{5}{c}{Forgery Methods ($\delta$)} \\ \cline{3-7}
~ & ~ & FaceSwap & Face2Face & FaceShifter & Deepfake & NeuralTexture \\ 

\hline
\hline

Source ($\phi_{v_s}$) & \multirow{2}{*}{ResNet-34 \cite{resnet}}
& 0.72 & 0.73 & 0.72 & 0.73 & 0.74 \\

Target ($\phi_{v_t}$) & ~ 
& 0.74 & 0.76 & 0.72 & 0.75 & 0.76 \\

\hline
\hline

Artifact ($\phi_{v_d}$) & ResNet-34 \cite{resnet}
& 0.34  & -0.02 & 0.18 & 0.00 & 0.04\\
\hline
\hline
Source ($\phi_{v_s}$) & \multirow{2}{*}{Efficient-b3 \cite{efficientnet}}
& 0.65 & 0.66 & 0.64 & 0.66 & 0.67 \\

Target ($\phi_{v_t}$) & ~ 
& 0.70 & 0.73 & 0.63 & 0.72 & 0.74 \\

\hline
\hline

Artifact ($\phi_{v_d}$) & Efficient-b3 \cite{efficientnet}
& 0.23 & 0.02 & 0.13 & -0.09 & -0.12 \\
\hline
\end{tabular}
\end{center}
\vspace{-4mm}
\end{table}

\section{More about the FST-Matching Deepfake Detection Model}
\subsection{Comparison with the baseline in terms of $\delta$.}
In this section, we compared the proposed metric $\delta$  between the baseline (\emph{i.e.} the detection encoder $v_d$) and the FST-Matching Deepfake Detection Model. 
Results in Table \ref{table:hypothesis3} show that compared to the baseline, artifact visual concepts learned by our model are more stable among compressed images.
Such results demonstrate the effectiveness of our method.
\begin{table}[h]
\setlength{\tabcolsep}{2pt}

\begin{center}
\caption{Comparison of the proposed metric $\delta$  between the baseline (\emph{i.e.} the detection encoder $v_d$) and the FST-Matching Deepfake Detection Model. 
The backbones of the baseline and our model are all ResNet-18 \cite{resnet}.
Results show that compared to the baseline, our model considers more similar visual concepts as artifact-relevant among compressed images.}
\linespread{1.3}
\scriptsize
\label{table:hypothesis3}
\begin{tabular}{l | c  c  c  c  c }
\hline
\multirow{2}{*}{Visual Concept} & \multicolumn{5}{c}{Forgery Methods ($\delta$)} \\ \cline{2-6}
~ & FaceSwap & Face2Face & FaceShifter & Deepfake & NeuralTexture \\ 
\hline
Artifact (Baseline)
& 0.17 & -0.02 & 0.14 & -0.15 & -0.14 \\
\hline
\hline
Artifact (FST-Matching)
& 0.54 & 0.46 & 0.47 & 0.45 & 0.40 \\
\hline
\end{tabular}
\end{center}
\end{table}

\subsection{Evaluation of the proposed metric $Q$.} 
Besides, we also evaluated the proposed FST-Matching Deepfake Detection Model via the proposed $Q$ to demonstrate its robustness to video compression. 
All models were trained on the raw dataset and tested on c23 and c40 compressed datasets afterwards.
We calculated the proposed metric $Q$ between the baseline model and our model.
The backbone of each model is set as ResNet-18 \cite{resnet}.
Results in Table \ref{table:FST-Q} show that our model has a significantly larger value of $Q$ on the c23 and c40 images, indicating the robustness of our model to different compression rates. 
Note that there exists a performance gap between the baseline and the FST-Matching Deepfake Detection Model on raw images. 
To this end, compared with the baseline in Table \ref{table:FST-Q}, our method is designed to explicitly disentangle the source/target-irrelevant representation from source/target visual concepts on images. 
Intuitively, such disentangled representation is less enriched than the overall representation of raw images learned by the baseline, causing the performance drop on raw images. 
However, the disentangled source/target-irrelevant representation is verified to be robust to video
compression in the paper, which facilitates our model to achieve great performance on compressed videos.

\begin{table}[h]
\setlength{\tabcolsep}{4pt}

\begin{center}
\caption{
Comparison of proposed metric $Q$ ($\times 10^{-2}$) between the baseline and the FST-Matching Deepfake Detection Model. 
Here $Q$ is averaged among different thresholds $\tau$ same as Table 1. 
Such results show that our model considers similar image regions as artifact-relevant visual concepts among different compressed images.
}
\linespread{1.6}
\tiny
\label{table:FST-Q}
\begin{tabular}{l | c  c | c  c | c  c }

\hline

\multirow{2}{*}{Forgery Methods} & \multicolumn{2}{c|}{Raw ($Q (\times 10^{-2})$)} & \multicolumn{2}{c|}{C23 ($Q (\times 10^{-2})$)} & \multicolumn{2}{c}{C40 ($Q (\times 10^{-2})$)} \\ \cline{2-7}

~ & \textit{Baseline} & \textit{FST-Matching} & \textit{Baseline} & \textit{FST-Matching} & \textit{Baseline} & \textit{FST-Matching} \\

\hline
\hline

FaceSwap \cite{faceswap}
& $\mathbf{2.77}$ & $2.08$ 
& $0.37 $ & $\mathbf{1.41}$ 
& -$0.52$ & $\mathbf{1.15}$ \\

Face2Face \cite{face2face}
& $\mathbf{2.31}$ & $1.97$ 
& -$0.87$ & $\mathbf{1.30}$ 
& -$1.35$ & $\mathbf{0.94}$ \\

FaceShifter \cite{faceshifter}
& $\mathbf{2.45}$ & $2.09$ 
& -$0.61$ & $\mathbf{1.05}$ 
& -$0.67$ & $\mathbf{0.72}$ \\

Deepfake \cite{deepfakes}
& $\mathbf{2.53}$ & $2.06$ 
& -$1.64$ & $\mathbf{1.22}$ 
& -$1.42$ & $\mathbf{0.87}$  \\

NeuralTexture \cite{neural-textural}
& $\mathbf{2.30}$ & $1.84$ 
& -$1.78$ & $\mathbf{0.71}$ 
& -$1.44$ & $\mathbf{0.05}$ \\
\hline
\end{tabular}
\end{center}
\end{table}

\clearpage
%
%

\bibliographystyle{splncs04}
\bibliography{fst-matching-camera-ready}

\end{document}